\documentclass{svproc}
\usepackage{url}
\usepackage{graphicx}
\usepackage{subcaption}
\begin{document}
\mainmatter              
\title{Study on Image Filtering - Techniques, Algorithm and Applications}
\titlerunning{Image Filtering}  
%
\author{Bhishman Desai\inst{1}
 \and Manish Paliwal\inst{2} and Kapil Kumar Nagwanshi* \inst{3}
}
\authorrunning{Desai et al.} 
%
\tocauthor{Bhishman Desai, Manish Paliwal, and Kapil Kumar Nagwanshi}
\institute{
Infogain, Mumbai, INDIA\\
\email{desaibhishman1@gmail.com},
\and
Pandit Deendayal Energy University, Raisan, 
Gandhinagar, Gujarat, INDIA \email{ paliwalmanish1@gmail.com}
\and
Department of Computer Science and
Engineering, \\ ASET, Amity University Rajasthan, Jaipur, India
\email{dr.kapil@ieee.org}\\
$^*$Corresponding Author
}

\maketitle              

\begin{abstract}
Image processing is one of the most immerging and widely growing techniques making it a lively research field. Image processing is converting an image to a digital format and then doing different operations on it, such as improving the image or extracting various valuable data. Image filtering is one of the fascinating applications of image processing. Image filtering is a technique for altering the size, shape, color, depth, smoothness, and other image properties. It alters the pixels of the image to transform it into the desired form using different types of graphical editing methods through graphic design and editing software. This paper introduces various image filtering techniques and their wide applications. 
\end{abstract}
\section{INTRODUCTION}

Image filtering such as smoothing an image reduces noise, blurred images can be rectified. There are broadly two types of algorithms- linear\cite{kailath1974view,satpathy2010image} and non-linear\cite{addabbo2019classification}. Linear filter is achieved through convolution and Fourier multiplication whereas Non-linear filter cannot be achieved through any of these. Its output is not the linear function of its input thus, its result varies in a non-intuitive manner. Here, the following image shows how the median filter\cite{juneja2009improved}  enhances the images by reducing the noise and smoothing. 
\begin{figure}[!htbp]
\begin{subfigure}[b]{0.5\textwidth}
         \centering
         \includegraphics[width=.5\textwidth]{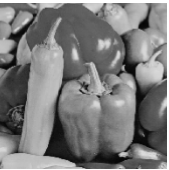}
         \caption{}
         \label{fig1a}
     \end{subfigure}
      \begin{subfigure}[b]{0.5\textwidth}
         \centering
         \includegraphics[width=.5\textwidth]{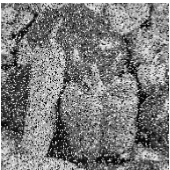}
         \caption{}
         \label{fig1b}
     \end{subfigure}
     \caption{(a)Original image , (b) Filtered image}

\end{figure} 

In order to do further processing like image segmentation, edge detection etc. noise should be eliminated. Median filter is the most effective non-linear filtering algorithm to detect and remove salt and pepper noise. Median filter retains the edges of the while removing the noise that's why it is used most widely thus, apart from trampling the noise up to 5\% to 60\%, it also preserves the image details. The noisy pixels are evaluated and labelled as noisy pixels and the switching based median filter is applied to other pixels which are not noisy. 

Bilateral filter \cite{paris2007gentle} is a type of non-liner filter, it reduces noise by smoothing and preserves edges of the images. It takes weighted sum of the pixels which are nearby of each pixel and replaces the intensity of the pixels with the average of that weighted sum. The neighbourhood pixels are identified through their locations which are relative to the input pixel. Algorithms \cite{chandel2013image} used for linear filtering are: Box blur, Gaussian, bilateral and Hann window.

\begin{figure}[!htbp]
    \centering
    \includegraphics[width=10cm]{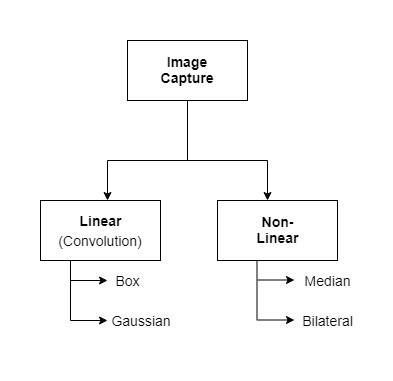}
    \caption{Types of Filtering Algorithms}
\end{figure}

In box blur \cite{zimmer2009blur}, an image with 9x9 pixel values, can be considered having a 3x3 neighbourhood values.  smoothing is achieved by averaging the neighbourhood pixel values of the particular pixel in the output image. In this way the pixels with higher intensity gets converted into a lower valued pixel and vice-versa thus, balancing the image pixels.
Gaussian filter is based on the equation of a Gaussian, it can be used to generate a kernel.
A kernel is a matrix (small) which is a convoluted matrix used in various image filtering techniques for embossing, smoothing, sharpening, blurring etc.
Gaussian blur can be applied for smoothing or filtering operations.

\subsection{Working Example- Box Blur:}

Let's take a 9*9 image as an input for the example. Here in the input image, 90 is the peak value and also some of the holes are 0 which are black point. Considering a 3*3 neighbourhood kernel around the pixel of which average values will be taken. Similarly, average values around all the pixels will be calculated by moving the kernel frame all over the input image to be able to generate newer versions of the output image. For example, if sum around a particular pixel is 360 then the value in corresponding pixel of output image will be 40 since, there are 9 pixels around the input pixel's kernel. Thus, in this manner all the values of the output matrix will be filled.

\begin{figure}[!htbp]
\begin{subfigure}[b]{0.5\textwidth}
         \centering
         \includegraphics[width=\textwidth]{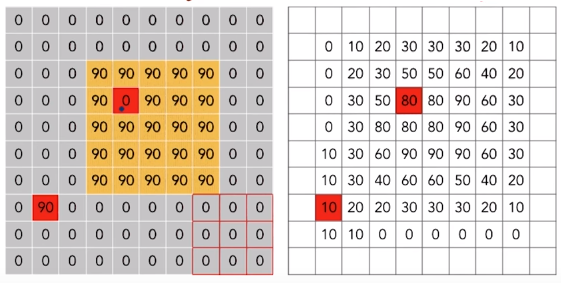}
         \caption{}
         \label{fig2a}
     \end{subfigure}
      \begin{subfigure}[b]{0.5\textwidth}
         \centering
         \includegraphics[width=\textwidth]{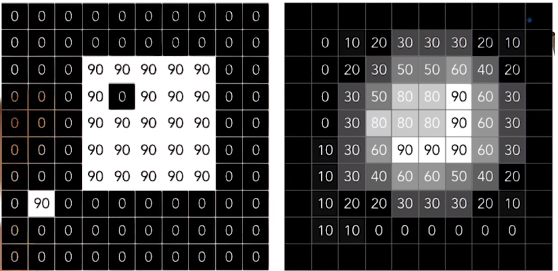}
         \caption{}
         \label{fig2b}
     \end{subfigure}
     \caption{(a)Input and  Output image , (b) Input and Output image in shades of gray}

\end{figure}

It can be observed from the image that the place where zero was surrounded by 90s, in the output image is replaced by a much higher value and thus smoothing intensity of the pixel. Also 90 which was surrounded by 0's is now reduced to 10 producing a smoothen output.

\section{Image filtering algorithms }
\subsection{Non-linear}
\subsubsection{Median Filter \cite{juneja2009improved}}
Before proceeding for filtering the image, noise should be detected properly in order to rectify the noise and smoother the image. Following assumptions are made for error detection i) in an image which is noise-free edges separate the smoothly varying areas, ii) the pixel which is noisy in the image has uneven intensity i.e. either very low or very high in comparison of the neighborhood pixels.
During the impulse noise detection procedure, two image sequences are generated. One is the series of gray-scale images,  $x^{(0)}_{(i,j)} , x^{(1)}_{(i,j)} ,\dots,x^{(n)}_{(i,j)}$.
Here, the first image $x^{(0)}_{(i,j)}$ is noisy image, ${(i,j)}$ is the position of pixels in the image where $1 \leq i \leq M, 1 \leq j \leq N$. Here, N and M represent direction of pixels in vertical and horizontal direction respectively and $x^{(n)}_{(i,j)}$ is the image after nth iteration. 

Noise detection procedure:
\begin{enumerate}
  \item	Assume a window $(2W+1) x (2W+1)$ around $x^{(n-1)}_{(i,j)}$  i.e. 
      \[ x^{(i+k,j+l)}_{(n-1)}\]  where $-W<=k<=W, -W<=l<=W$ and $W>=1$.

  \item	Finding windows, median value $m^{(i,j)}_{(n-1)}$  ,
        \[m^{(i,j)}_{(n-1)}   = median (x^{(i+k,j+l)}_{(n-1)})\]
  \item	The absolute difference of $x^{(n-1)}_{(i,j)}$   and  $m^{(i,j)}_{(n-1)}$  ,
\[f^{(i,j)}_{(n)} =
\begin{cases}
      {f ^{(i,j)}_{(n-1)},   if | x^{(n-1)}_{(i,j)}  -   m^{(i,j)}_{(n-1)} | <T}\\
      {1, otherwise}
    \end{cases} \]
\item	If noise is detected in (i,j)th  then the value of x(n)(I,j)   will be changed as,
\[x^{(n)}_{(i,j)} =       
\begin{cases}
     m^{(i,j)}_{(n-1)}  ,    if f^{(i,j)}_{(n)} \neq f^{(i,j)}_{(n-1)} \\
     x(n-1)(i,j)  ,    if f^{(i,j)}_{(n)} \neq f^{(i,j)}_{(n-1)}
    \end{cases} \]  

\end{enumerate}
All these steps needed to be repeated for p times. Where p can be 2,3,4…. and so on.
After the pth iteration, $x^{(p)}_{(i,j)}$  and  $f^{(p)}_{(i,j)}$    are two images formed. But for noise filtering process, the binary flag image $f^{(p)}_{(i,j)}$ is only required. Sun and Neuvo introduced this median based noise-based detection and later Wang and Zhang (PSM) modified it in a progressive manner. This method or approach is distinct from PSM as in this method various adaptive filtering approach as it focuses and improves the filter's filtering performance. 

\subsubsection{Bilateral Filtering}
   It is a simple and non-iterative method. Bilateral filtering\cite{weiss2006fast} preserves the edges while smoothing the noisy image by virtue of combination of non-linear nearby imagee value. Based on photometric similarity and geometric closeness. In this method, no phantom colours along the edges of the output images are produced. 
    Bilateral filter is similar to Gaussian convolution as it is the average of the pixels but the bilateral filter considers the variations in intensities thus, preserves the edges of the image. Bilateral filtering can be justified as when the two pixels are near to each other and they are similar in terms of the photometric range along with the occupancies of the spatial locations which are nearby.
\subsection{Gaussian \textit{vs.} Bilateral}
The heat equation or isotropic diffusion has a Gaussian correlation. It applies uniformly to the picture, thereby encompassing the borders and image content. The Gaussian and Bilateral filters are nearly identical, except that the Gaussian inflects through a modulated function that determines the distance between the center and neighboring pixels. There is no filtering at the object's boundaries with Gaussian filtering, and the filter is applied evenly throughout the picture. On the other hand, a bilateral filter prevents blurring between objects while yet being successful in evenly eliminating noise and so giving a superior outcome.

\section{Applications}
\begin{enumerate}

\item \textit{Image Enhancement:} Image filtering helps in enhancing the quality of the image pixels. It comprises of the processes of altering the images in terms of pixel values (blurring, smoothing etc.) irrespective of the image type i.e. photochemical, digital or even illustration, this processing is known as photo retouching in case of the analog image enhancing. Various softwares are also used which encompasses three dimensional modelers, vector and raster graphic editors. These softwares are primary tools through which an image can be enhanced and transformed as desired\cite{jain1989fundamentals}. \\
\begin{figure}[!htbp]
    \centering
    \includegraphics[width=8cm]{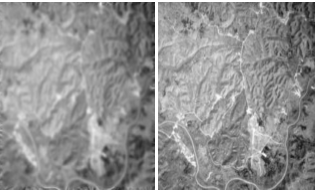}
    \caption{Noisy Image and Filtered Image}
\end{figure}

\item \textit{Denoising:} A ten function is used as a range weight, while a square box function is used as a spatial weight. The bilateral filter, unlike Gaussian blur, delivers sharper and superior results by focusing on maintaining the object outlines. It has many applications in fields like movie restoration and medical imaging.
\begin{figure}[!htbp]
\begin{subfigure}[b]{0.33\textwidth}
         \centering
         \includegraphics[width=.8\textwidth]{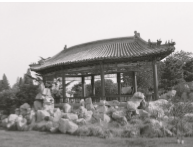}
         \caption{}
         \label{fig3a}
     \end{subfigure}%
      \begin{subfigure}[b]{0.33\textwidth}
         \centering
         \includegraphics[width=.8\textwidth]{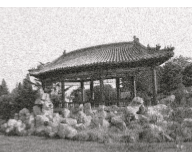}
         \caption{}
         \label{fig3b}
     \end{subfigure}%
     \begin{subfigure}[b]{0.33\textwidth}
         \centering
         \includegraphics[width=.8\textwidth]{den-noisy}
         \caption{}
         \label{fig3c}
     \end{subfigure}
     \caption{(a)Input image , (b) Noisy image, (c) Surface blur filter}

\end{figure} 

\item \textit{Image Compression:}
Compression can lead to partial data loss i.e., it leads to lossy compression. But in the field of medical image processing \cite{schalkoff1989digital}, clipart, comics, and technical drawings, lossless image data is preferred. Compression artifacts can be introduced during lossy compression methods when short-bit rates are used. The compression, which is lossy but doesn't create unobtrusiveness, can be called visually lossless.

\item \textit{Texture and illumination separation:} Texture and illumination separation is done in image based modelling. The uneven texture in images is removed by the bilateral filter successfully and also preserves the cessation which are caused by geometrical and brightness changes. This method is inspired from the fact that in an image, variations in brightness are much higher as compared to that in texture patterns\cite{katsaggelos2012digital,satpathy2010adaptive,satpathy2010image}.

\begin{figure}[!htbp]
    \centering
    \includegraphics[width=8cm]{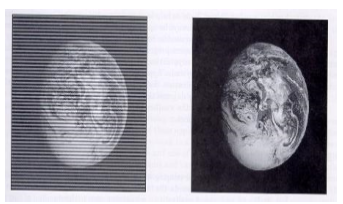}
    \caption{Noisy Image and Restored Image}
\end{figure}

\item \textit{Character Recognition:} It is a common procedure to digitize printed compositions with the ultimate goal that they can be altered, looked at and stored even more firmly in machine procedures, For instance, webbased machine interpretation, content todiscourse, extraction of important data and content mining. OCR is a knoll of insight, example, and PC vision research.Early shapes needed to be computerized with each character's images and worked on one text style at a time. There are some appealing techniques for copying engineered yields, particularly after the first filtered sheet, including sections, images and other unprinted segments\cite{mollah2011design}.

\item \textit{Signature Verification:} A computerized mark is a science scheme to speak to an advanced correspondence's authenticity.A lawful computerized mark bears the beneficiary's motivation to think that a perceived sender made the message to such an extent that the sender could not refuse to send the message with nonrepudiation and confirmation and that the message was not changed in exchange.Advanced marks are usually used to define programming, budget correspondence, and in other circumstances where impersonation or alteration is essential.

\item \textit{Biometrics:}
To depict the last class of biometrics, some researchers have instituted the word behavior measurements \cite{nagwanshi2012b,nagwanshi2013d,nagwanshi2018s,nagwanshi2019c}. The enmity of Fingerprint Verification (FVC) is an intercontinental challenge focused on evaluating the programming of distinctive mark confirmation. A subgroup of distinctive mark impressions was provided to enlisted people with different sensors allowing the parameters of distinct calculations to be changed. Individuals were spoken to offer enlist and match executable records of their calculations ; the evaluation was carried out at the courtesies of the coordinators using the submitted compelling documents on an appropriate database acquired as the preparation set with the indistinguishable sensors \cite{nagwanshi2020estimation,basheer2021fesd,nagwanshi2012introduction}.

\item \textit{Object Recognition:} Item discovery is a PC innovation recognized with PC vision and image preparation that manages to see depictions of semantic objects of a class, such as individuals, constructions or cars in computerized recordings and images. Well-inquired regions of discovery of article integrate face location and recognition of walkers. Item recognition is guaranteed in many areas of PC vision, similar to picture recovery and video observation.

\begin{figure}[!htbp]
\begin{subfigure}[b]{0.5\textwidth}
         \centering
         \includegraphics[width=\textwidth]{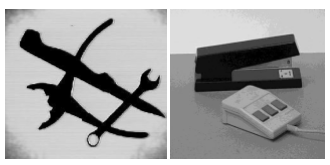}
         \caption{}
         \label{fig2a}
     \end{subfigure}
      \begin{subfigure}[b]{0.5\textwidth}
         \centering
         \includegraphics[width=\textwidth]{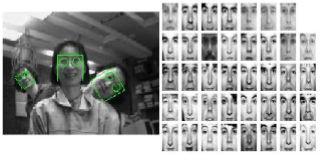}
         \caption{}
         \label{fig2b}
     \end{subfigure}
     \caption{(a)Images in Object Recognition , (b) Face Images.}
\end{figure} 

\item \textit{Face Detection:}
Face identification is subject to PC technology that constructs nonsensical (sophisticated) photos of the dimensions and regions of human faces \cite{ahmad2013image,satpathy2010image}. 
\item \textit{Medical Applications:} In particular, we attach importance to restorative imagery as the critical use of image handling. Despite the fact that imaging can be performed for medicinal subtleties of expelled organs and tissues, such occasions are not usually referred to as therapeutic imaging, but to some extent they are a piece of pathology.As a reprimand and in its broadest context, it is a piece of natural imaging and includes radiology that utilizes Xbeam radiography imaging advances, appealing reverberation imaging, ultrasonographic or ultrasound therapeutic material, endoscopy, elastography, imaging, thermography, restorative and nuclear prescription helpful imaging techniques as positron outflow tomography\cite{Raja21,mahmood19,patra20}.
\end{enumerate}

\section{Conclusions}
Image filtering has a wide range of applications including edge detection, removing noise, sharpening and smoothing. Image filtering has brought an evolutionary change in the field of image processing.  A filter can be characterized as a kernel, which can be defined as a small array applied to every pixel and their adjacent neighbors in the given image. Here, in this paper we have briefly explained about the image filtering and compared various algorithms. We have also briefly discussed about its various applications such as image enhancement, image compression and restoration, denoising and in medical field. When it is needed to reduce noise while preserving the peaks and edges of the image, median and bilateral filter is recommended. But, if only the peaks of the image are needed to be preserved and effect on the edge is not to be bothered then gaussian filter is sufficient as it requires less computation comparatively. Whereas if only noise reduction is concerned and peaks and edges are not in consideration, box filter will suffice. It is recommended keeping the kernel size less than or equal to 7 for any algorithm for optimal computational cost. 

\bibliography{sample}
\end{document}